\label{key}
\documentclass[letterpaper, 10 pt, journal, twoside]{IEEEtran}
\usepackage[linesnumbered,ruled]{algorithm2e}
\usepackage{array}
\usepackage[caption=false,font=footnotesize,labelfont=rm,textfont=rm]{subfig}
\usepackage{stfloats}
\usepackage{url}
\usepackage{verbatim}
\usepackage{booktabs}
\usepackage{makecell}
\usepackage{multirow}

\usepackage{cite}
\usepackage{amsmath,amssymb,amsfonts}
\usepackage{algorithmic}
\usepackage{graphicx}
\usepackage{makecell}
\usepackage{textcomp}
\usepackage{enumerate}
\usepackage{bm}
\usepackage{color}
\usepackage{framed}
\usepackage{xcolor}
\usepackage{epstopdf}

\makeatletter
\renewcommand{\maketag@@@}[1]{\hbox{\m@th\normalsize\normalfont#1}}%
\makeatother

\usepackage{hyperref} 
\begin{document}

\title{Contact SLAM: An Active Tactile Exploration Policy Based on Physical Reasoning Utilized in Robotic Fine Blind Manipulation Tasks}

\author{Gaozhao Wang, Xing Liu$^{*}$, Zhenduo Ye, Zhengxiong Liu and Panfeng Huang
	\thanks{Gaozhao Wang, Xing Liu(Corresponding author), Zhenduo Ye, Zhengxiong Liu and Panfeng Huang
		are with the Research Center for Intelligent Robotics, the National Key
		Laboratory of Aerospace Flight Dynamics, and Shaanxi Province Innovation Team
		of Intelligent Robotic Technology, School of Astronautics, Northwestern
		Polytechnical University, Xi'an 710072, China. (e-mail: gaozhao\underline{ }wang@mail.nwpu.edu.cn.)}
}



\maketitle

\begin{abstract}
	Contact-rich manipulation is difficult for robots to execute and requires accurate perception of the environment. In some scenarios, vision is occluded. The robot can then no longer obtain real-time scene state information through visual feedback. This is called ``blind manipulation". In this manuscript, a novel physically-driven contact cognition method, called ``Contact SLAM", is proposed. It estimates the state of the environment and achieves manipulation using only tactile sensing and prior knowledge of the scene. To maximize exploration efficiency, this manuscript also designs an active exploration policy. The policy gradually reduces uncertainties in the manipulation scene. The experimental results demonstrated the effectiveness and accuracy of the proposed method in several contact-rich tasks, including the difficult and delicate socket assembly task and block-pushing task.
\end{abstract}


\section{Introduction}
In robotic manipulation tasks, vision plays an irreplaceable role. Through
visual servoing, a robot can identify the target of manipulation and control
its manipulator to complete the task \cite{marcos2025map}. In this process, the
tactile or force sensing is mainly used to interact with the environment after
the vision has already provided the target position \cite{zhang2025vtla}.

\begin{figure}[ht]
	\centerline{\includegraphics[width=8.5cm]{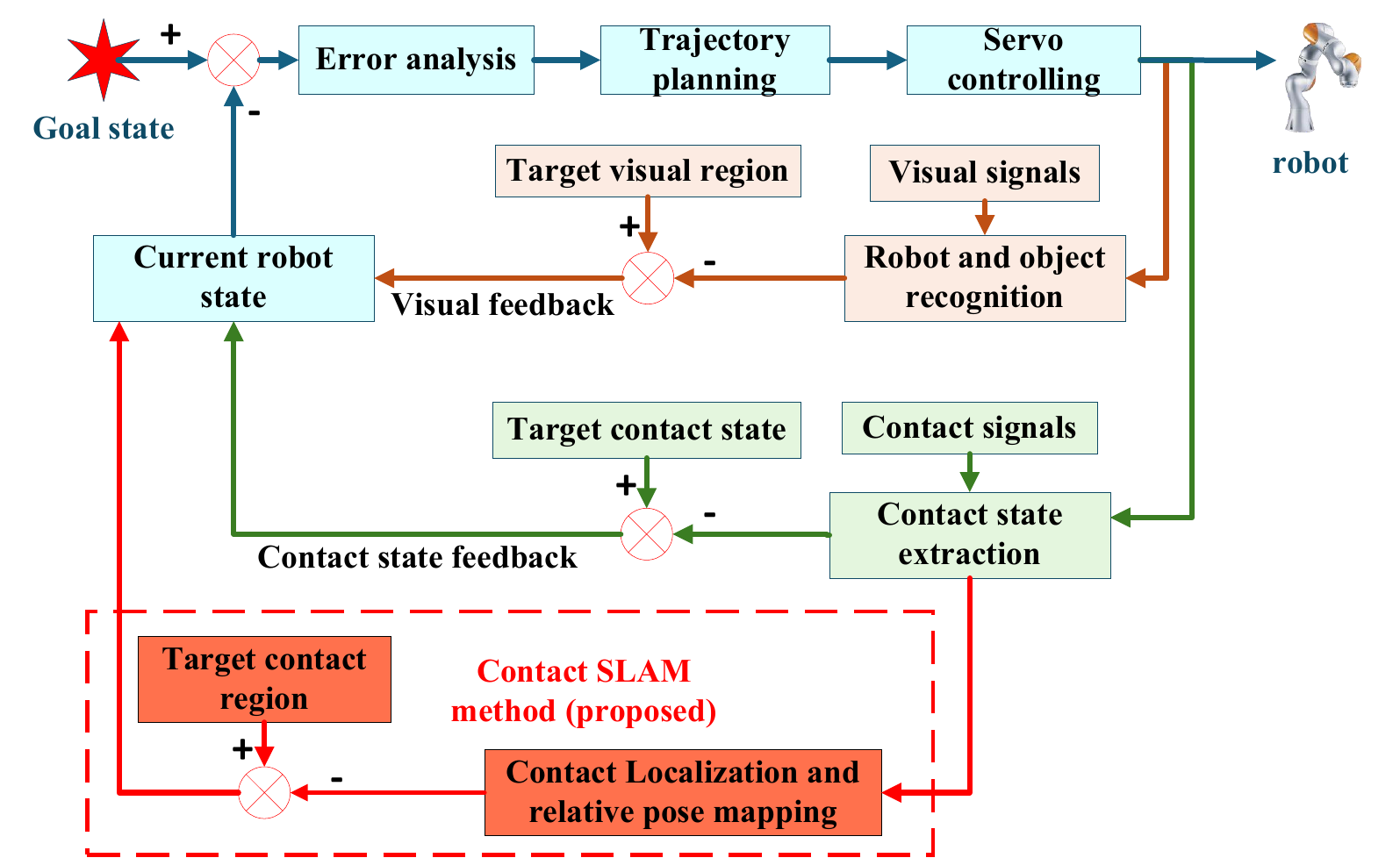}}
	\caption{The perception and cognition process in robot manipulation tasks. In general, the visual signals are used to understand the scene. When the vision is occluded, other kinds of information, such as tactile signals, are used to obtain contact states. However, the contact state is hard to be labelled and classified. We proposed the contact SLAM method, which doesn't pursue obtaining an exact contact state. Instead, it acquires the contact area with the environment through the cognitive process during the exploration of the environment.}
	\label{problem_motivation}
\end{figure}

However, in certain tasks, when vision fails or is occluded, the robot can no
longer obtain real-time scene state information via visual feedback. We refer
to such manipulation tasks as “blind manipulation” tasks, in which the robot
must rely solely on tactile and force information. Compared with visual
sensing, tactile sensing provides more comprehensive information about the
contact state with the environment during manipulation, including the geometry
of the contacted object\cite{xu2023tandem3d}, the contact force/torque and its
distribution\cite{zhang2019effective}, and contact event
triggering\cite{yu2018realtime}. Some researches have tried to transform the
force or tactile signals into some forms of contact state
\cite{migimatsu2022symbolic, lee2025vitascope, Cao2024Uncertain}, The
manipulation based on contact states tends to focus on low-level state feedback
and control, such as determining whether specific signals relevant to task
completion are present during contact\cite{noseworthy2025forge} or designing
contact feedback control loops\cite{bi2025vla, xue2025reactive, dutta2023push}
to replicate particular contact signals, but these methods usually have strong
data dependency, and the utilized scene is constrained strictly. With these
perceptual features, they can be further applied to robotic manipulation
tasks\cite{bauza2023tac2pose, zhao2023skill}. In terms of interaction with the
environment, end-to-end policy generation is typically implemented through a
sim-to-real approach\cite{bauza2024simple}. Although some studies have
investigated the perception process during manipulation with tactile sensing,
they typically stop at the perception stage without conducting an in-depth
study on task completion. The deatiled comparison between the visual and
tactile servo control process is shown in Fig. \ref{problem_motivation}

During human manipulation, in addition to visual awareness of the process,
there is also contact awareness. By judging the force sensed at the fingertips,
people can estimate whether the manipulation goal has been reached and, if not,
how far they are from the target state. This is a typical simultaneously
localization and mapping(SLAM) problem. Sudharshan Suresh et
al.\cite{suresh2021tactile} proposed using force–tactile sensing to
simultaneously track an object’s position and model its contour, a method they
termed Tactile SLAM. Jialiang Zhao et al.\cite{zhao2023fingerslam} introduced
the concept of FingerSLAM. Paloma Sodhi et al.\cite{sodhi2021learning} applied
the SLAM modeling concept to object manipulation control, achieving
tactile-based pushing operations. However, in blind manipulation tasks, except
for the missed viisual information, there also exists uncertainties, which
could prevent the task to be finished, but the existing tactile methods don't
consider this.

In manipulation planning process, the uncertainty eliminating is widely used in
the contact process\cite{saleem2025contact, saleem2024pomdp}, but they usually
use the binary force singal, and they often cosider one target object without
interference from other objects. So Inspired by tactile SLAM methods and
contact-driven blind manipulation methods, we introduce physically-driven
contact cognition into robotic blind manipulation. By reasoning about the
contact state between the grasped object and the environment from contact
signals, the robot can continuously estimate and explore the environment until
ultimately completing the manipulation task, as illustrated in
Fig.\ref{problem_motivation}.

The contributions of this paper are as follows:

\begin{enumerate}
	\item{\textbf{A contact state-based pipeline for executing the contact-rich manipulation is proposed.} The contact-rich manipulation process is defined as the contact state translation process, and the relative pose between the grasped object and scene could be inferred from the contact state.}
	\item{\textbf{A contact event-based SLAM method for environment perception and localization is proposed}. This method defines the contact event and further develops and extends the scope of tactile SLAM, proposing a post-estimation framework for relative pose estimation, enabling the perception and understanding of the scene state based on prior scene knowledge and contact event feedback.}
	\item{\textbf{An effective planning method for fine blind manipulation based on active tactile exploration is proposed}. This approach balances environmental exploration with task completion efficiency, plans optimal action strategies according to the current understanding of the environment, and reduces the uncertainties of the manipulation scene gradually.}
\end{enumerate}

%
%
%
%
%
%

\section{Problem Formulation: Contact SLAM Analysis}\label{problem_formulation}
In blind manipulation tasks, several kinds of objects need to be noticed. The
grasped objects, which are grasped by the gripper, and the tactile sensors
could only perceive the tactile signals between the grasped objects and the
gripper. The manipulated objects, which are used to interact with the grasped
objects, including unmovable obstacles and movable objects. The last is the
target region, which the grasped objects or manipulated objects need to reach.
The detailed scenes are shown in section \ref{experimental_setup}.

\begin{figure}[ht]
	\centerline{\includegraphics[width=8cm]{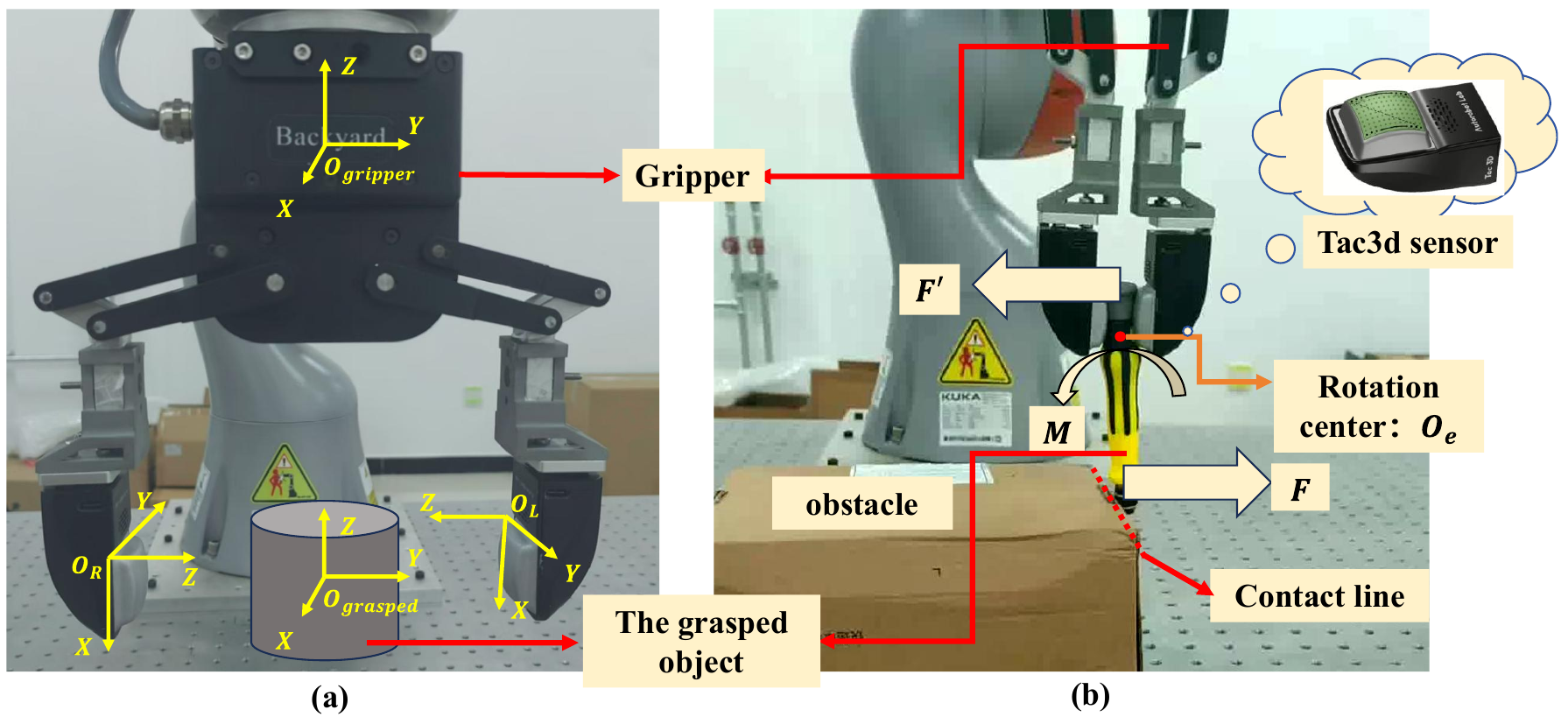}}
	\caption{(a) represents the coordinates used in this manuscript. The left and right sensor coordinates are opposite to the left and right positions in space because of sensor setting. (b) represents the corresponding forces and torques applied to the object when contact happens.}
	\label{coordinate_description}
\end{figure}

In order to represent the pose and of these things, the coordinates used in
this manuscript are illustrated in Fig. \ref{coordinate_description}(a), and
the iteration between the grasped object and obstacles is shown in Fig.
\ref{coordinate_description}(b).

To achieve the task objectives, we make the following assumptions:
\begin{enumerate}
	\item{There exists a relative motion tendency between the grasped object and the gripper, but no relative translational displacement.}
	\item{The contact between objects is quasi-static.}
	\item {The precise geometric dimensions and shapes of the grasped object and the manipulated object are known in advance.}
	\item {All objects in the environment are static and do not move autonomously; the only moving bodies are the gripper and the objects it holds.}
\end{enumerate}

Contact SLAM aims to address the problems of relative pose localization between
objects and environment reconstruction in contact-rich manipulation tasks, and
then plans the subsequent end-effector motion trajectory for the robot.

According to the definition of SLAM, contact SLAM also includes both
localization and mapping components:

\begin{itemize}
	\item[$\bullet$] \textbf{Localization} focuses on determining the positions of the robot end-effector, the grasped object, the target object, and other objects of interest in the environment.

	\item[$\bullet$] \textbf{Mapping} involves determining the positions of the landmarks in the environment by representing and analyzing the current contact information between the grasped object and the environment. This approach does not seek to obtain an accurate contact configuration of the grasped object within the environment; instead, it only requires relative contact-region information, which is sufficient to satisfy the manipulation task requirements.
\end{itemize}

\begin{figure*}[ht]
	\centerline{\includegraphics[width=16.5cm]{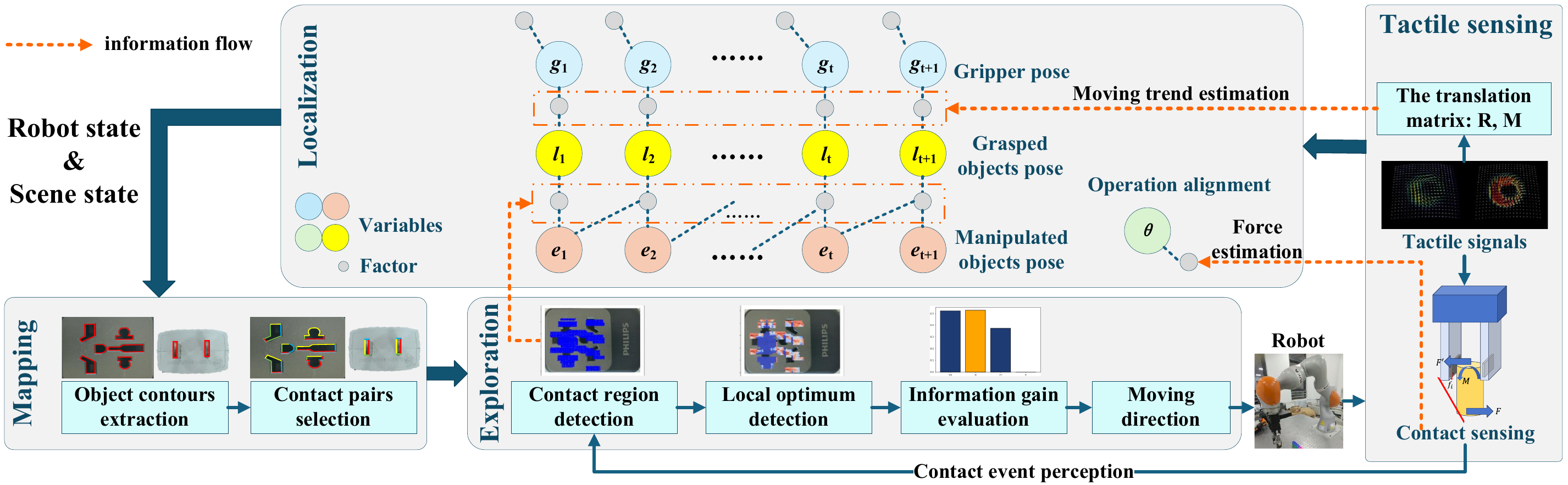}}
	\caption{The complete process of the proposed method. The method regards the manipulation process as an SLAM problem. In the localization phase, it focuses on determining the positions of the robot end-effector, the grasped object, the target object, and other objects of interest in the environment. In the mapping phase, it focuses on representing the current relative contact information together with the tactile information between the grasped object and the environment. After one SLAM process, generate the action strategy based on the information gain evaluation in different movement directions, then the robot executes the strategy, the tactile sensing converts the tactile signals into physical information and sends them to the SLAM process.}
	\label{contact_slam_flow}
\end{figure*}

The complete contact SLAM method includes tactile perception, localization,
mapping and active exploration process, as illustrated in Fig.
\ref{contact_slam_flow}.

\section{Methods}
\subsection{Contact Force Estimation Analysis }\label{tactile_perception}
The Tac3D sensor used in this study provides the three-dimensional force
distribution, the resultant three-dimensional force, and the resultant
three-dimensional torque in the sensor coordinate frame. These enable
researchers to determine not only the magnitude but also the trend of forces
acting on the grasped object.

When a two-finger gripper grasps an object, all forces acting on the grasped
object are reflected in the tactile sensors mounted on the gripper fingers. The
two-finger gripper and the grasped object could be modeled as a beam fixed at
one end. When an external force is applied to one end, the reaction at the
gripper includes not only a force in the opposite direction but also a torque
relative to the grasping point. Both the forces and torques are applied to the
tactile sensors mounted on the gripper.

As illustrated in Fig. \ref{coordinate_description}(b), when a force in a
certain direction acts on the grasped object, it could be decomposed into three
components along the $X$, $Y$, and $Z$ axes of the gripper coordinate system.
To maintain the stability of the grasped object, the two gripper fingers
produce corresponding torques on the object. The relationship between the
forces applied to the object and those measured by the gripper tactile sensors
can be expressed as follows:

\begin{small}
	\begin{equation}\label{eq1}
		\left\{\begin{matrix}
			F_x=F_y^R-F_y^L \\
			F_y=F_z^L-F_z^R \\
			F_z=F_x^L-F_x^R \\
		\end{matrix}\right.
	\end{equation}
\end{small}
where $F_x, F_y$, and $F_z$ are force components in the frame of the gripper. $F_i^L$ and $F_i^R$ are corresponding force components ($i=x, y, z$) in the frame of the left sensor and right sensor.

\subsection{State Perception and Optimization}\label{contact_slam}
We employ factor graph optimization as our estimation framework, which is based
on the principle of maximum a posteriori (MAP) estimation. A factor graph is a
bipartite graph with two types of paprmeters: variables $x$ and factors $\phi$.
Variable nodes are the latent states to be estimated, and factor nodes encode
constraints on the variables, such as measurement likelihood functions, or
physics, geometric models\cite{sodhi2021learning}. Under Gaussian noise model
assumptions, MAP inference is equivalent to solving a nonlinear least-squares
problem. That is, for Gaussian factors $\phi_i(x)$ corrupted by zero-mean,
normally distributed noise, the inference equation is:

\begin{small}
	\begin{equation}\label{eq7}
		\hat{x}=\underset{x}{\mathrm{argmin}}\frac{1}{2}\sum_{i=1}^{n}{\left \| F_i(x) \right \|}^2_{{\sum}_i}
	\end{equation}
\end{small}

In contact SLAM, following factors are optimized:

\begin{itemize}
	\item[$\bullet$] \textbf{Gripper Localization Factor $ F_{gri}$}: The position of the gripper $g_t$ can be derived from the position of the robot’s end-effector:

	      \begin{small}
		      \begin{equation}\label{eq8}
			      {\left \| F_{gri}(g_t) \right \|}^2_{{\sum}_{gri}}={\left \| g_t-g_t^{pri} \right \|}^2_{{\sum}_{gri}}
		      \end{equation}
	      \end{small}

	\item[$\bullet$] \textbf{Grasped object Localization Factor $ F_{obj}$}: The position of the object grasped by the gripper $l_t$ can be obtained from the gripper position and the relative displacement of the object sensed by the tactile sensor mounted on the gripper. Based on the prior method\cite{motion_perception_icra_2021}, we define several coordinate frames: the gripper coordinate frame $(OXYZ)_{g}$, sensor coordinate frame $(OXYZ)_{s}$, grasped object coordinate frame $(OXYZ)_{o}$, and world coordinate frame $(OXYZ)_{w}$. Our estimation target is the position of a grasped object in the world coordinate frame, and the factor expression is:

	      \begin{small}
		      \begin{equation}\label{eq9}
			      {\left \| F_{obj}(l_t, g_t) \right \|}^2_{{\sum}_{obj}}={\left \| l_t-T^w_gT^g_sT^s_lI \right \|}^2_{{\sum}_{obj}}
		      \end{equation}
	      \end{small}
	      here, $T^w_g$ represents the gripper pose which could be derived from the robot’s end-effector pose, $T^g_s$ is fixed during operation, and $T^s_l$ depends on the relative displacement of the marker points sensed by the tactile sensor, which could be decomposed as: $T^s_t=T^s_{sd}T^{sd}_l$, where $T^{sd}_l$ is unchanged. To obtain $T^s_{sd}$, the SVD-based optimization proces is acquired:

	      \begin{small}
		      \begin{equation}\label{eq10}
			      R^\ast, M^\ast =\min_{R,M}\sum_{i=1}^{N}\left \| p_i-(R\cdot p_i^{\prime}+M) \right \| ^2
		      \end{equation}
	      \end{small}
	      where the translation matrix $T^s_{sd}$ is composed of the optimized rotation matrix $R^\ast$ and translation matrix $M^\ast$.

	\item[$\bullet$] \textbf{Environment Localization Factor $F_{env}$}: Based on the contact pairs obtained through active exploration and the object position at time t, the position of obstacles $e_t$ can be estimated. The estimated pose of environment in $(OXYZ)_w$ is:

	      \begin{small}
		      \begin{equation}\label{eq11}
			      \begin{aligned}
				      \hspace{-2mm}
				      \begin{split}
					       & {\left \| F_{env}(e_t, e_{t-1}, l_t) \right \|}^2_{{\sum}_{env}}= \\&{\left \| B(l_t + B(P_t)) \cup B(l_t-B(P_t)) \cap e_{t-1} - e_t \right \|}^2_{{\sum}_{env}}
				      \end{split}
			      \end{aligned}
		      \end{equation}
	      \end{small}
	      where $B(\cdot)$  means the boundary of the objects or obstacles' position. The parameter $P_t$ is the distribution range of particles at the current time, which is shown in section \ref{active_exploration}. Initially, the $e_0$ is set to a range that could cover the potential position of interested objects.

	\item[$\bullet$] \textbf{Operation Alignment Factor $F_{ali}$}: During manipulation, researchers are not concerned with the absolute position of the object being manipulated; rather, they focus on whether the correct contact state has been achieved. Therefore, we define the following operation alignment factor:

	      \begin{small}
		      \begin{equation}\label{eq13}
			      {\left \| F_{ali}(\theta_{ali}) \right \|}^2_{{\sum}_{ali}}={\left \| E(F_x \cap F_y \cap d)-\theta_{ali} \right \|}^2_{{\sum}_{ali}}
		      \end{equation}
	      \end{small}
	      where $F_x, F_y$ are force components calculated by eqution \ref{eq1}, and $d$ represents the distance. $E(\cdot)$ represents a function used for measuring the error between the current state and the goal state. $\theta_{ali}$ is a binary independent variable, if $\theta_{ali}=1$, the task is considered to be finished.

\end{itemize}

\subsection{Contact Localization and Active Tactile Exploration Policy}\label{active_exploration}
In the process of blind manipulation, the exact pose of the object is unknown.
To achieve the goal of estimating relative contact regions, the system must
generate exploratory actions to interact with the environment, continuously
estimate the contact state during the contact process, and adjust the action
based on the tactile sensor feedback. Based on these, we draw inspiration from
active localization methods\cite{agarwal2015motion} and propose the
\textbf{Contact Localization and Active Tactile Exploration Policy} method. The
detailed workflow is shown as follows:

\textbf{Preparation stage:} At this stage, given that the exact geometric shapes of the objects in the environment and the grasped object are all represented as polygons, and their vertices $v_1, v_2,..., v_N$ are collected in counterclockwise order. The contour edges can then be computed as: $e_i=v_{i+1}-v_i$, and the outward normal vector of each contour edge is given by: $n_i=(e_i^y, -e_i^x)$. Thus, we can define the contour of the grasped object and contours of other objects in the environment: $S_{e_i}=\{(n_i, v_i, v_{i+1})|i=1,...,N_i\}$, $S_{l_j}=\{(n_j, v_j, v_{j+1})|j=1,...,M_j\}$. And $C_{env}=\{S_{e_1}, S_{e_2},...,S_{e_N}\}$, $C_{object}=\{S_{l_1}, S_{l_2},...,S_{l_M}\}$.

After that, to represent the relative position state between the grasped object
and the scene objects, the distribution of the reference point within the
environment is initialized. In the manuscript, we choose the particle filtering
method. The distribution of particles at time $t$ is represented as $P_t$, and
the weight of each particle is represented as $w_i^t=1/len(P_t)$.

\textbf{Step 1, Local optimum detection.} Examine the weights and choose the local optimum as follows: $peaks=\{p_i^t|p_i^t \in P_t, w_i^t>0.5/len(P_t), i=1,2,...n\}$. If the boundary measurement of peaks within $\delta_{thr}$, the distribution of peaks is taken as the potential region of the reference point. If not, the process turns to Step 2 to select a movement direction.

\textbf{Step 2, Information gain evaluation:} For each potential motion direction of the object, the expected information gain is calculated, which is represented as the types of contact states of motion distance. We choose the entropy as the justance criterion, and calculate the entropy of each action as follows:

\begin{small}
	\begin{equation}\label{eq15}
		entropy(Z_a)=\sum_{z \in Z_a} -p(z) \log p(z)
	\end{equation}
\end{small}
Where $Z_a$ is the predicted contact state set of all particles with the action $a$.

Afer getting the information gain of each potential action, we choose the
action which could get the most information as the moving direction:
$\max_a\{entropy(Z_a)\}$.

\textbf{Step 3, Exploration through motion:} The grasped object moved in the selected direction until a change in the tactile signal is detected.

\textbf{Step 4, Contact pair selection and updation of particles:} After detecting a tactile signal change, we can estimate the direction of the force acting on the grasped object. From the perspective of object dynamics, the force direction and the outward normal vector of the object’s contour satisfy the following relationship: $F_G^t \cap n_i >0, F_G^t \cap n_j <0$. Thus, the contact pairs can be represented as: $(i,j)=\{n_i=-n_j\}$, and the distribution of particels is updated as follows: $P_t=P_{t-1} \cap (B(S_{e_i}) \cup B(S_{l_j}))$.




\textbf{Step 5, Particles' weights updation:} After selecting the candidate contact pairs, we then backtrack the contact events during the motion trajectory. Specifically, based on the traveled distance $Distance$ and the number of motion steps $T$, we infer the potential contact states at intermediate time steps. If at any of these time steps, the inferred contact state is inconsistent with the observed tactile signal, the weight of the corresponding particle is reduced. This process is repeated until the weights of all particles are updated. The weight $w_i^t$ is updated as follows:

\begin{small}
	\begin{equation}\label{eq14}
		w_i^{t+1}=w_i^t \cdot \mathcal{P}(z_{obs}^t,z_{pred}(p_i - (T-t)/T \cdot Distance))
	\end{equation}
\end{small}
where $z_{obs}^t$ represents the tactile obserbation at the $t$ time, $z_{pred}$ represents the predicted tactile signal at the pose of $p_i - (T-t)/T \cdot Distance$.

After updating the weights of particles, the local optimum detection process is
repeated, and the particles are further updated: $P_t=peaks$.

If particles' distribution is less than the predefined threshold, the process
turns to Step 7 to find the optical policy to finish the task. If not, the
process turns to Step 6 to resample more particles.

\textbf{Step 6, Resample particles:} After updating the weights of all particles, we check the number of particles. If the number of particles is less than a predefined threshold (e.g., 50), we resample additional particles within the current potential region to ensure sufficient particle density for accurate localization. Based on the particle filter method, we choose to resample the same number of particles each iteration.

Return to the Step 1 and continue the exploration process.

\textbf{Step 7, The generation of final manipulation policy: }When the particles' distribution is smaller than the set threshold, the manipulation policy is generated. The optimal policy is represented as moving from the estimated object pose to the target pose. In the motion planning process, we choose the A\_star algorithm as the planning method.

\subsection{The improved tips of Active Exploration}\label{improve_tips}
In order to improve the efficiency of the active exploration, we adopt the
following improvements:


First, the distance variance criterion is added to the process of calculating
the information gain. When the distribution of particles is small, the entropy
calculation could become very small, too. That means the information gain is
useless in the choice of action direction. When the entropy calculation is
small, the distance variance is added to the information gain. A larger
distance criterion value indicates greater variability in the motion of the
current particle distribution, enabling more particles to be distinguished
based on the discrepancy between the estimated motion distance and the actual
motion distance.

The deatils of Gain information calculation is shown in Algorithm \ref{alg1}.

\renewcommand{\algorithmicrequire}{\textbf{Input:}}
\renewcommand{\algorithmicensure}{\textbf{Output:}}
\begin{algorithm}[!h]\label{alg1}
	\caption{Information Gain Evaluation}
	\begin{algorithmic}[1]
		\REQUIRE
		\STATE The potential distribution of movable objects: $P_t=peaks$;
		\STATE The referance actions:$A=\{a_1, a_2,...a_m\}$;
		\STATE The contour of environment: $ C_{env}=\{S_{e_1}, S_{e_2},..., S_{e_N}\} $\;
		\STATE The contour of grasped object: $ C_{object}=\{S_{l_1}, S_{l_2},..., S_{l_M}\}$\;

		\ENSURE
		\IF{$len(P_t)=1$:}
		\STATE Choose the optical trajectory;
		\ENDIF
		\IF{$len(P_t)>1$:}
		\FOR{$a \in A$}
		\FOR{$p \in P_t$}
		\STATE Calculate the contact state: $z\_pred$ at $p$ with action $a$;
		\STATE Add $z\_pred$ to $Z_a$;
		\STATE Calculate the distance $d$ to get the state $z\_pred$;
		\STATE Add $ d$ to $D_a$;
		\ENDFOR
		\STATE Calculate the entropy of $Z_a$: $entropy(Z_a)=\sum_{z \in Z_a} -p(z) \log p(z)$;
		\STATE Calculate the variance of $D_a$: $var(D_a)=\sum_{d \in D_a} (d - \mu(D_a))^2$;
		\ENDFOR
		\IF {$\max \{entropy(Z_a)\} < e_{thr}$:}
		\STATE Choose the action based on the Distance variance: $\max_a\{var(D_a)\}$;
		\ELSE
		\STATE Choose the action which could get the most information: $\max_a\{entropy(Z_a)\}$.
		\ENDIF
		\ENDIF
	\end{algorithmic}
\end{algorithm}

Second, we adjust the number of particles to resample. Instead of resampling
the same particles in each episode. In environments with imperfect or imprecise
modeling, overly rapid particle depletion may cause some plausible distribution
regions to be discarded prematurely. A straightforward solution is to increase
the number of particles during resampling. However, if the resampled particle
count is fixed, an insufficient number of particles fails to improve
localization accuracy, while an excessive number leads to particle overcrowding
in later iterations, or even to a continuous increase in the total particle
count during the filtering process. To balance exploration efficiency and
localization accuracy, we therefore design an adaptive particle resampling
mechanism. This mechanism maintains a reasonable particle density in all
potentially valid distribution regions, thereby reducing the probability that
correct hypotheses are eliminated due to incidental or unexpected events.

The whole \textbf{Active Tactile Exploration policy} is shown in Algorithm
\ref{alg3}.

\renewcommand{\algorithmicrequire}{\textbf{Preparation:}}
\renewcommand{\algorithmicensure}{\textbf{Output:}}
\begin{algorithm}[!h]\label{alg3}
	\caption{Active Tactile Exploration Policy}
	\begin{algorithmic}[1]
		\REQUIRE
		\STATE Scene construction and obtain $C_{env}, C_{object}$;
		\STATE Initialization of the reference point distribution: $P_t=\{p_1, p_2,...,p_n\}$, $W_t=\{w_1^t, w_2^t,..., w_n^t\}$;
		\ENSURE
		\STATE Local optimum detection: $P_t=peaks=\{p_i^t|p_i^t \in P_t, w_i^t>1/len(P_t), i=1,2,...n\}$;
		\STATE Information gain evaluation and obtain action $\pi$;
		\STATE Execute $\pi$ and monitor force until the contact occurs;
		\STATE Updation of particles: $P_{t+1}=P_{t} \cap (B(S_{e_i}) \cup B(S_{l_j}))$;
		\STATE Update the particles' weights: $w_i^{t+1}=w_i^t \cdot \mathcal{P}(z_{obs}^t,z_{pred}(p_i - (T-t)/T \cdot Distance))$, $W_{t+1}={w_i^{t+1}|i=1,2,...,len(P_{t+1})}$;
		\STATE Resample particles: $P_{t+1}^{\prime}=AR(P_{t+1}, P_{t}, W_{t+1}) $
		\IF {$B(peaks)>\delta_{thr}$}
		\STATE return to ``Local optimum detection".
		\ELSE
		\STATE Find the optimal trajectory $\pi$;
		\STATE Execute $\pi$, until the alignment factor satisfies $\theta_{ali}=1$;
		\ENDIF
	\end{algorithmic}
\end{algorithm}

\section{Experimental Studies}
\subsection{Experimental Setup}\label{experimental_setup}
\begin{figure}[ht]
	\centerline{\includegraphics[width=7cm]{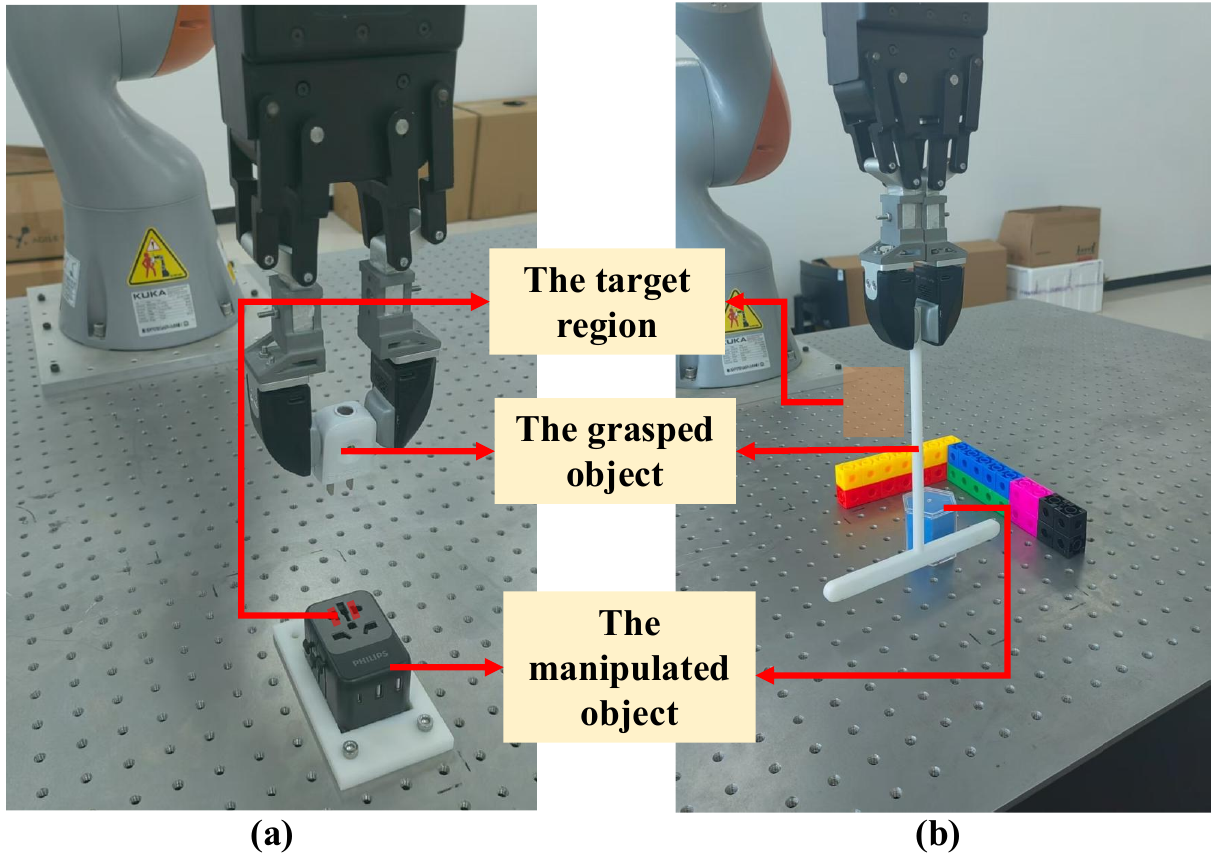}}
	\caption{The calibration task scenario and two manipulation experiment scenarios. (a) represents the calibration process of the Tac3D sensor; (b) represents the socket assembly task scenario; (c) represents the block-pushing task scenario.}
	\label{tactile_manipulation_exp}
\end{figure}

This manuscript defines two blind manipulation task scenarios, which are
illustrated in Fig. \ref{tactile_manipulation_exp}.

The first task scenario involves a blind peg-in-hole assembly of a power socket
component of two different standards. The robot is required to grasp the socket
and insert it into the correct receptacle. The second task scenario is a blind
pushing task in the presence of obstacles. In this scenario, the operator has
prior knowledge of the approximate position of the block to be pushed, the
target location, as well as the geometrical shapes of the obstacles present in
the environment. However, the exact positions of these obstacles within the
scene are unknown. The chosen two tasks are all typical blind indirect contact
manipulation tasks, because the contact with the scene needs to be referred
from the contact signals between the gripper and the grasped object.

\begin{itemize}
	\item[$\bullet$] \textbf{Evaluation metrics:} After the mapping process, we calculate the localization error represented as root mean square error(RMSE) and iteration times. After the manipulation process, we evaluate the whole exploration time of manipulation, as well as the success rate.

	\item[$\bullet$] \textbf{Compared methods:} We choose the \textbf{Spiral-hole-searching} method, which is a typical passive searching method, as the baseline and control method.

\end{itemize}

\subsection{Results of the Socket Assembly Experiments}
In this experiment, we used two different plugs: a two-pin plug and a
triangular three-pin plug, to assemble into the same socket, as illustrated in
Fig. \ref{tactile_manipulation_exp}(b). The main challenge of this task lies in
the need to distinguish between different assembly regions corresponding to
different plug geometries. In the absence of visual guidance, the robot must
rely solely on tactile feedback and prior knowledge of the scene to identify
and reason about the relative contact regions between the plug and the socket.

\subsubsection{The ablation experiments}
We introduce two improvements to the particle filtering procedure, namely a
distance variance criterion and adaptive resampling. The necessity of these
improvements is demonstrated through ablation experiments. Specifically, for
each type of socket pin, we select a random initial pose, and conduct the
control group(fixed resampling without the distance variance criterion) and two
experimental groups, one using the distance criterion alone and the other
combining the distance criterion with adaptive resampling.

\begin{figure*}[ht]
	\centerline{\includegraphics[width=16cm]{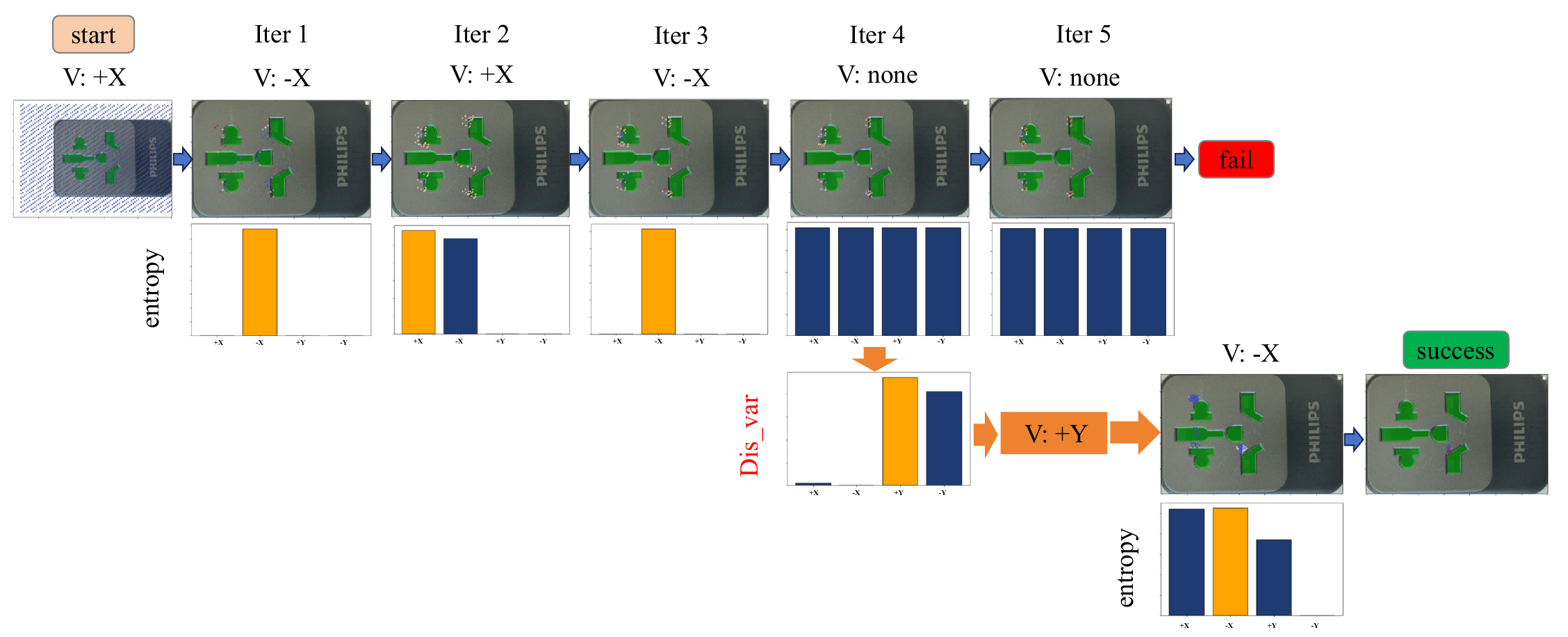}}
	\caption{The comparison of the control group and experimental group adding the distance criterion. In iteration 4, the entropy criterion is small and unique in all directions, indicating that the active exploration failed. After introducing the distance criterion, the optimal information gain is obtained.}
	\label{comparation_of_Disvar}
\end{figure*}

For the control group, since entropy is the only information criterion, when
the particle distribution becomes highly concentrated, the resulting motion
direction is unique in all cases, as shown in Fig. \ref{comparation_of_Disvar}.
This indicates that the entropy criterion fails under such conditions. To
address this issue, we introduce the \textbf{distance variance criterion} as
shown in Section \ref{improve_tips}. The specific implementation is as follows:
when all entropy values fall below $10^{-2}$, the decision criterion is
switched from entropy to the distance criterion. We then reattempt exploration
in the previously failed scenarios. As shown in Fig.
\ref{comparation_of_Disvar}, the results demonstrate that, after introducing
the distance criterion, effective environment exploration can still be achieved
even when the entropy criterion is extremely small, thereby leading to correct
localization results.

\begin{figure*}[ht]
	\centerline{\includegraphics[width=14cm]{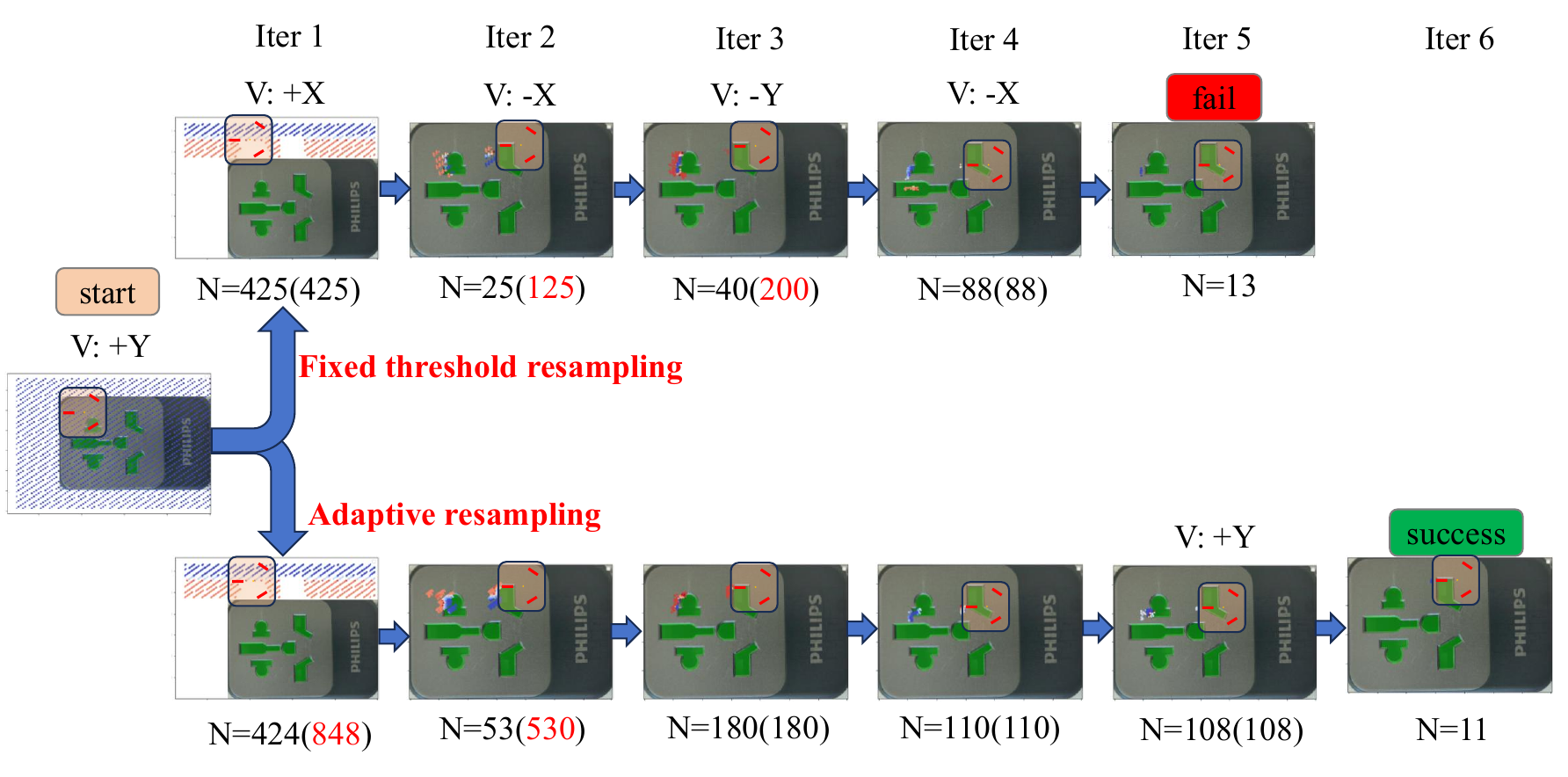}}
	\caption{The comparison of the control group and experimental group utilizes adaptive resampling. In iteration 5, the particle set around the plug was eliminated because of the accumulated position errors during exploration. In the experimental group, }
	\label{comparation_of_Adresample}
\end{figure*}

In the control group, a resampling strategy was adopted in which resampling is
triggered only when the number of particles falls below 50. Under this
strategy, accumulated particle position errors in earlier stages may result in
certain particle modes being eliminated from the new particle set before the
resampling threshold is reached, as illustrated in Fig.
\ref{comparation_of_Adresample}, ultimately causing localization failure. We
added the \textbf{adaptive resampling} to the filter process as shown in
Section \ref{improve_tips}, and the rule of resampling is as follows:

\begin{eqnarray}\label{eq16}
scale & = & \left\{\begin{matrix}
 \frac{P_t}{3 \times P_{t+1}} & len(P_{t+1})>300\\
  \frac{P_t}{2 \times P_{t+1}} & 100\le len(P_{t+1}) \le 300 \\
  \frac{P_t}{P_{t+1}} & 50\le len(P_{t+1}) < 100 \\
  \frac{2 \times P_t}{P_{t+1}} & len(P_{t+1}) < 50 
\end{matrix}\right.
\end{eqnarray}

\begin{eqnarray}\label{eq17}
 P_{t+1}^{\prime} = scale \times P_{t+1}
\end{eqnarray}

The results are shown in Fig. \ref{comparation_of_Adresample}. It can be
observed that, after incorporating adaptive resampling, the particle
distribution was better maintained, and the localization process successfully
converged to a unimodal distribution.

\subsubsection{The comparative experiments}
To verify the accuracy of the proposed method for estimating the relative
position of the peg and the hole, we conducted 25 sets of experiments for each
pin assembly task. The initial position of each set of experiments was randomly
selected within the plane of the hole. In total, 50 sets of experiments were
carried out for data analysis. For the control group, we also conducted 25 sets
of experiments for each pin at arbitrary initial positions, making a total of
50 sets of experiments for comparison. For the experimental group, when the
estimation of the relative position between the plug and the socket converges
to a unimodal distribution, we use the method as described in the article
\cite{morgan2023towards} for the final assembly operation. For the spiral
hole-searching method used for comparison, we set the following threshold
conditions: $F_{x/y} > 0.55*F_z$ if $ F_z > 0.5$; $F_{x/y} > 0.9 $ if $ F_z <
	0.5$. If the threshold conditions are met, it is considered that the hole area
has been found, and then the plug is inserted downward and it is checked
whether the assembly is completed. For both the experimental group and the
control group, the conditions for successful assembly are: $d_z > 0.025$ and
$F_{x/y} > 0.9N$.

\begin{table}[ht]
	\centering
	\caption{The results of socket assembly experiments.}
	\renewcommand{\arraystretch}{0.9}
	\label{tab1}
	\scalebox{1.0}{
		\begin{tabular}{ccccc}
			\hline
			              & \textbf{RMSE} & \textbf{Iter.Num} & \textbf{Explor.time} & \textbf{SR} \\
			\hline
			two-pin       & $ 0.449mm $   & $ 7.70 $          & $ 507.31s $          & $80\%$      \\

			three-pin     & $ 0.698mm $   & $ 9.95 $          & $ 679.96s $          & $84\%$      \\

			sprial search & --            & --                & $ 165.04s$           & $10\%$      \\ \hline
		\end{tabular}}
\end{table}

The experimental results are shown in Tab. \ref{tab1}. It can be seen that the
success rate of assembly using the contact positioning and active exploration
tactile method is significantly higher than that of the control group. For the
assembly tasks of the two types of bolts we selected, the success rates reached
$80\%$ and $84\%$ respectively. In contrast, the spiral hole-searching method
used for comparison only succeeded once in all attempts. This indicates that
compared with the passive hole-searching method, the hole-searching method with
active reasoning ability has significant advantages when facing an environment
with redundancy and interference. The hole-searching method with active
reasoning ability can model the environmental uncertainty, determine the
possible potential positions, and then further narrow down the possible
potential positions during continuous interaction, gradually eliminating the
uncertainty. On the other hand, the passive search method is only suitable for
situations with small errors. When there are large errors and environmental
interference, due to the limitations of observation, it is likely to miss the
correct position and fall into local optima.

\begin{figure*}[ht]
	\centerline{\includegraphics[width=15cm]{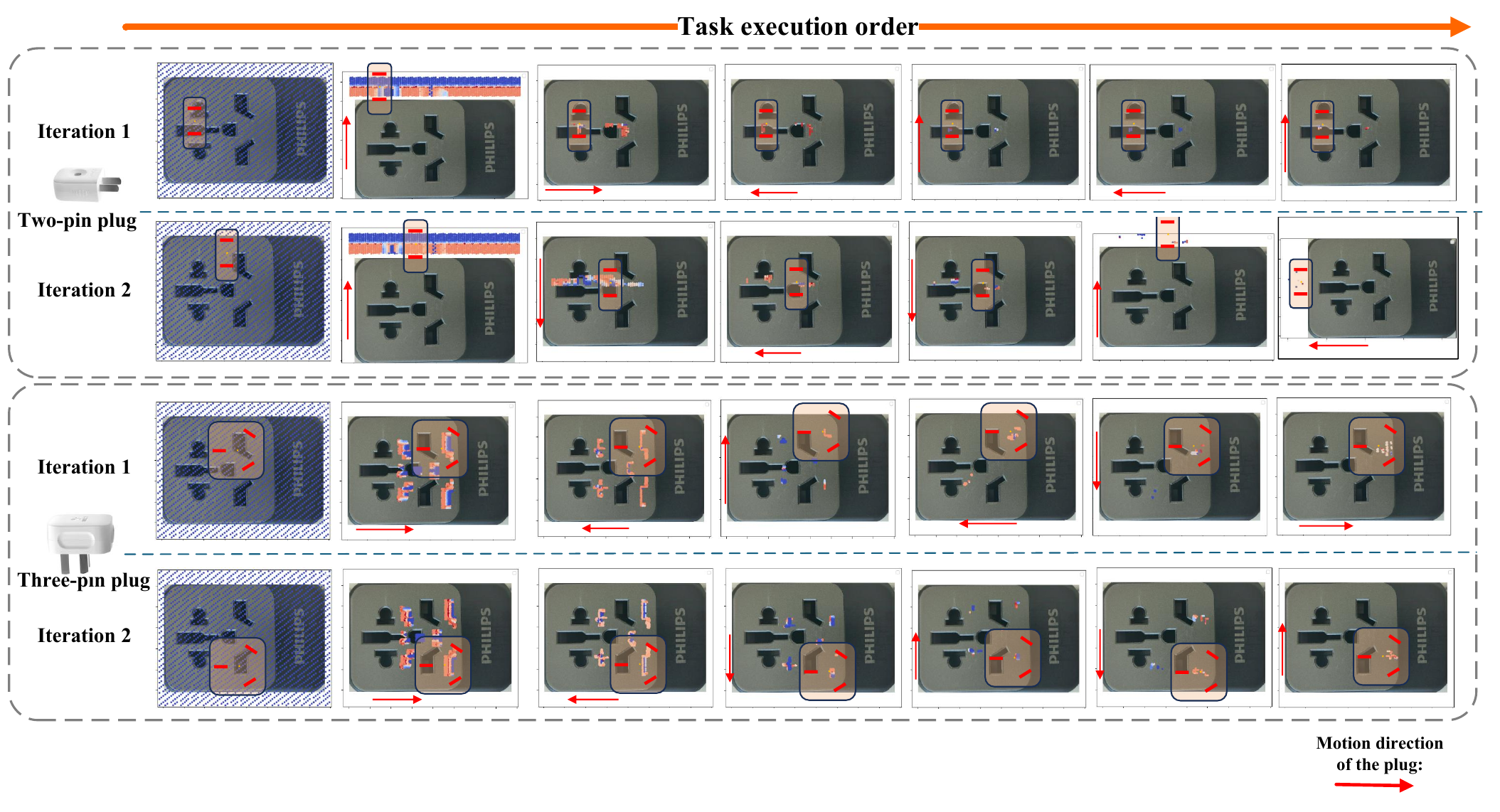}}
	\caption{The results of socket assembly experiments in different initial locations. The distribution of particles represents the potential relative pose of the plug and the socket. Red particles represent that the plug could be located at this point with more possibility, while the blue particles represent the less possibility.}
	\label{socket_assembly_res}
\end{figure*}

Fig. \ref{socket_assembly_res} intuitively shows the process of relative
position positioning using contact-based positioning and active tactile
exploration. It can be seen that when the edge of the plug collides with the
edge of the hole or goes beyond the hole area, the positions and weights of the
particles are redistributed. Particles with too low weights or outside the
potential distribution area are deleted. After $7 \sim 9$ explorations, the
distribution of the particles gradually stabilizes. If there is a single-peak
distribution at this time, it indicates that the exploration is successful.
Failed cases are mainly concentrated in situations where the active exploration
fails to converge, that is, there is still a multi-peak distribution after the
maximum number of exploration steps. The reason for the non - convergence of
the exploration is that the hole area is complex and the sizes of different
areas are similar. If the plug falls into a hole area with a similar size, the
exploration movement range will be greatly limited. If the plug has not
explored the overall environmental area comprehensively at this time, the
particles will remain in the hole areas with similar sizes and cannot converge
to a single-peak distribution.

\begin{figure}[ht]
	\centerline{\includegraphics[width=8.8cm]{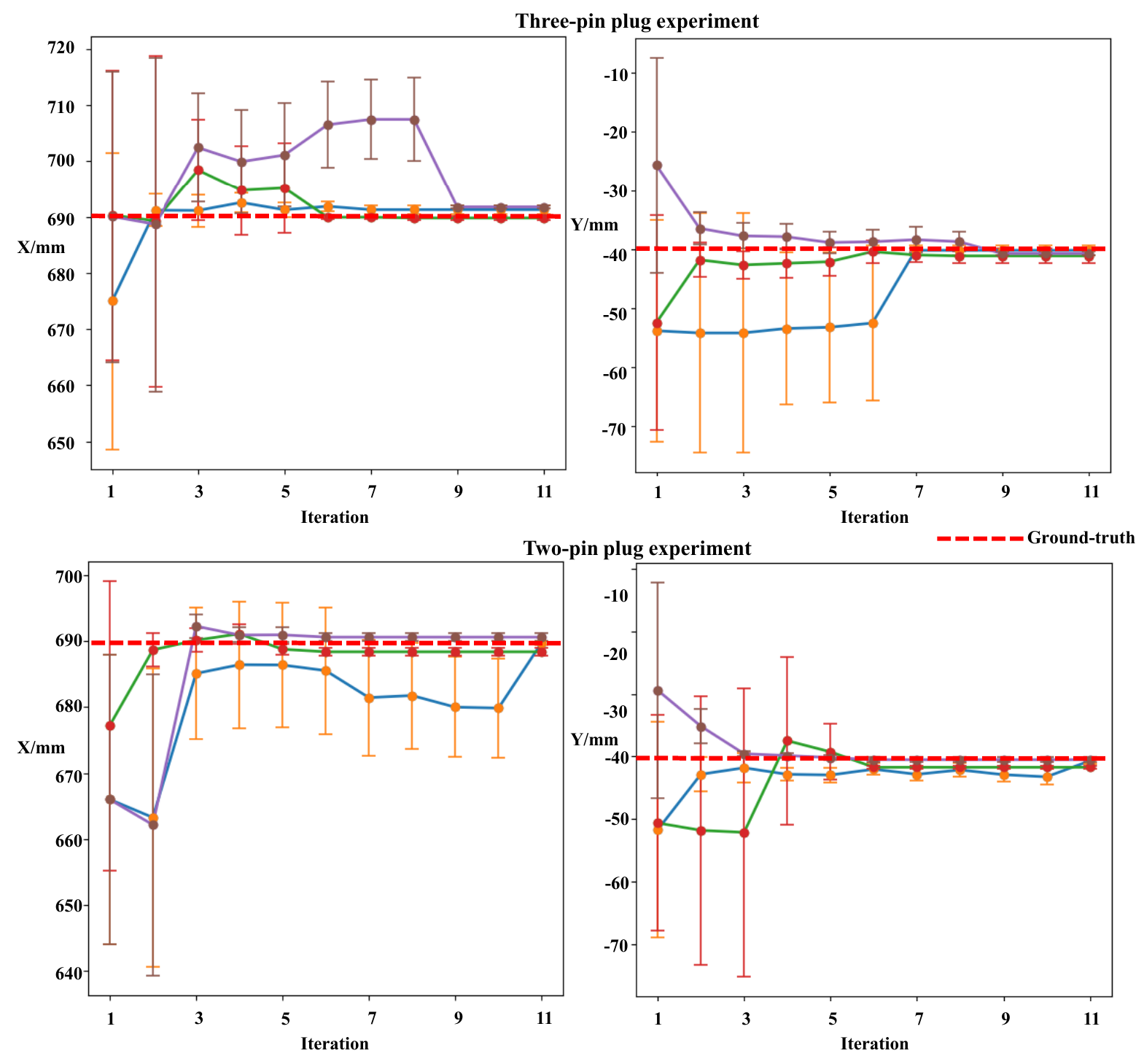}}
	\caption{The estimation results of the socket. Different initial position of gripper and plug got the same pose estimation of socket after several iterations.}
	\label{localization_test}
\end{figure}

To verify the accuracy of the position localization results of contact SLAM, we
verified the position estimation results of the socket obtained during the
active exploration process. The results are shown in Fig.
\ref{localization_test}.

The reason why we chose to estimate the socket position in the environment as a
verification of the positioning accuracy of contact SLAM is that during the
task execution process, the positions of the gripper and the grasped object are
constantly changing, and only the environment remains unchanged. As can be seen
from Fig. \ref{localization_test}, with the convergence of the particle
distribution, the results of socket position estimation starting from different
initial positions will converge to the vicinity of the true value after
exploration, and the error is within 2 mm, indicating the effectiveness of our
contact SLAM for unknown estimation.


\subsection{Results of the Blind Pushing Experiments}\label{relative_position_verification}
\begin{figure*}[ht]
	\centerline{\includegraphics[width=16.5cm]{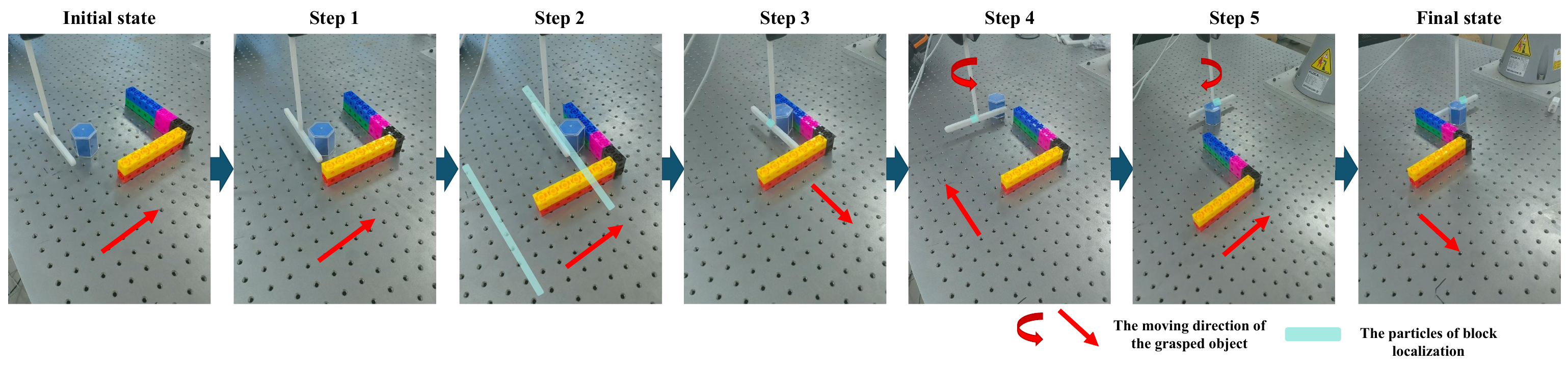}}
	\caption{The detailed results of the block-pushing task. The particle distribution represents the potential position ranges of the block and the obstacles.}
	\label{pushing_block_res}
\end{figure*}

In this task, the robot must first detect the contact region between the block
and the T-shaped tool. When obstacles are encountered, the robot estimates
their relative positions, replans the task trajectory accordingly, and
ultimately pushes the block into the designated target region. The experimental
results are illustrated in Fig. \ref{pushing_block_res}.

As shown in Fig. \ref{pushing_block_res}, in this task scenario, when obstacles are
encountered, the robot is able to reduce the localization error of the
obstacles to within 10 mm through a limited number of active explorations and
contact interactions, which is sufficient to meet the requirements of
trajectory re-planning.

\subsection{Experimental Results Discussion}
In the socket assembly experiment scenario, the success of perception depends
on the complexity of the contact between the plug and the socket. For a two-pin
plug, there are two planes in contact with the socket plane. Although the
geometric shape of a three-pin plug is more complex, the part in contact with
the socket plane is only its longest prong. Therefore, the contact situation of
the two-pin plug is more complex, which also affects the number of exploration
attempts. The number of exploration attempts for the two-pin plug is higher
than that for the three-pin plug.

In the blind block-pushing experiment, it is crucial to distinguish between the
movable block and the immovable obstacles. When an obstacle is encountered, the
analysis of the particle distribution must take into account the size of the
movable block. Compared with the case where the grasped object is in direct
contact with the obstacle, the resulting particle distribution exhibits a
translation shift.

\section{Conclusion}
This manuscript proposed a novel framework named contact SLAM, which could be
utilized for fine blind manipulation tasks. The framework integrates tactile
perception and physical reasoning to enable scene understanding and motion
planning. Given prior knowledge of the objects’ geometrical shapes and
dimensions, it allows the robot to precisely localize the relative pose between
the grasped object and the manipulated object solely through tactile sensing,
thereby accomplishing manipulation tasks without relying on visual information.
The authors validated the effectiveness of the proposed framework through
experiments on peg-in-hole assembly and block-pushing tasks. The present study
primarily focuses on geometric constraints. In future work, the authors plan to
extend this approach to incorporate dynamic constraints in fine manipulation,
aiming to further enhance the robot’s capability in executing such tasks.


\bibliographystyle{IEEEtran}
\bibliography{active_haptic_exploration_control}\

\vfill

\end{document}